\documentclass[10pt,letterpaper]{article}
\usepackage{lmodern}
\usepackage{amssymb,amsmath}
\usepackage{ifxetex,ifluatex}
\usepackage{fixltx2e} 
\ifnum 0\ifxetex 1\fi\ifluatex 1\fi=0 
  \usepackage[T1]{fontenc}
  \usepackage[utf8]{inputenc}
\else 
  \ifxetex
    \usepackage{mathspec}
  \else
    \usepackage{fontspec}
  \fi
  \defaultfontfeatures{Ligatures=TeX,Scale=MatchLowercase}
  
\fi
\IfFileExists{upquote.sty}{\usepackage{upquote}}{}
\IfFileExists{microtype.sty}{%
\usepackage{microtype}
\UseMicrotypeSet[protrusion]{basicmath} 
}{}
\usepackage[top=0.85in,left=2.75in,footskip=0.75in]{geometry}
\usepackage{hyperref}
\PassOptionsToPackage{usenames,dvipsnames}{color} 
\hypersetup{unicode=true,
            pdfborder={0 0 0},
            breaklinks=true}
\usepackage{graphicx,grffile}
\makeatletter
\def\maxwidth{\ifdim\Gin@nat@width>\linewidth\linewidth\else\Gin@nat@width\fi}
\def\maxheight{\ifdim\Gin@nat@height>\textheight\textheight\else\Gin@nat@height\fi}
\makeatother
\setkeys{Gin}{width=\maxwidth,height=\maxheight,keepaspectratio}
\providecommand{\tightlist}{%
  \setlength{\itemsep}{0pt}\setlength{\parskip}{0pt}}
\let\rmarkdownfootnote\footnote%
\def\footnote{\protect\rmarkdownfootnote}

\usepackage{titling}

  \title{}
  \pretitle{\vspace{\droptitle}}
  \posttitle{}
  \author{}
  \preauthor{}\postauthor{}
  \date{}
  \predate{}\postdate{}

\usepackage[nomarkers, nolists, figuresonly]{endfloat}
\usepackage{amsmath,amssymb}
\usepackage{changepage}
\usepackage{textcomp,marvosym}
\usepackage{cite}
\usepackage{nameref,hyperref}
\usepackage{microtype}
\DisableLigatures[f]{encoding = *, family = * }
\usepackage[table]{xcolor}
\usepackage{array}
\newcolumntype{+}{!{\vrule width 2pt}}
\newlength\savedwidth

\usepackage{setspace}
\raggedright
\usepackage[aboveskip=1pt,labelfont=bf,labelsep=period,justification=raggedright,singlelinecheck=off]{caption}

\makeatletter
\renewcommand{\@biblabel}[1]{\quad#1.}
\makeatother
\def\bx{{\bf x}}
\def\by{{\bf y}}

\def\thick#1{\hbox{\rlap{$#1$}\kern0.25pt\rlap{$#1$}\kern0.25pt$#1$}}

\def\bgamma{{\thick\gamma}}

\def\btheta{{\thick\theta}}

\def\brho{{\thick\rho}}

\def\bpsi{{\thick\psi}}

\def\smbalpha{{\thick{\scriptstyle{\alpha}}}}

\def\smbalpha{\widehat{\smbalpha}}

\def\hbar{{\overline h}}

\def\underrightarrow{\mathpalette\underrightarrow@}
\def\underrightarrow@#1#2{\vtop{\ialign{##\crcr$\m@th\hfil#1#2\hfil$\crcr
\noalign{\nointerlineskip}\rightarrowfill@#1\crcr}}}

\def\etal{{\em et al.}}
\def\lboxit#1{\vbox{\hrule\hbox{\vrule\kern6pt
      \vbox{\kern6pt#1\kern6pt}\kern6pt\vrule}\hrule}}

\def\thickboxit#1{\vbox{{\hrule height 1mm}\hbox{{\vrule width 1mm}\kern6pt
          \vbox{\kern6pt#1\kern6pt}\kern6pt{\vrule width 1mm}}
               {\hrule height 1mm}}}

\def\fat#1{\hbox{\rlap{$#1$}\kern0.25pt\rlap{$#1$}\kern0.25pt$#1$}}

\usepackage{booktabs}

\usepackage{lastpage,fancyhdr,graphicx}
\usepackage{epstopdf}
\fancyheadoffset[L]{2.25in}
\fancyfootoffset[L]{2.25in}

\ifx\paragraph\undefined\else
\let\oldparagraph\paragraph
\renewcommand{\paragraph}[1]{\oldparagraph{#1}\mbox{}}
\fi
\ifx\subparagraph\undefined\else
\let\oldsubparagraph\subparagraph
\renewcommand{\subparagraph}[1]{\oldsubparagraph{#1}\mbox{}}
\fi

\begin{document}

\vspace*{0.2in}

\begin{flushleft}
{\Large
\textbf\newline{Prediction of infectious disease epidemics via weighted density ensembles} 
}
\newline
\\
Evan L. Ray\textsuperscript{*},
Nicholas G. Reich
\\
\bigskip
Department of Biostatistics and Epidemiology, University of Massachusetts, Amherst, MA, USA
\\

\bigskip

* elray@umass.edu

\end{flushleft}

\section{Abstract}\label{abstract}

Accurate and reliable predictions of infectious disease dynamics can be
valuable to public health organizations that plan interventions to
decrease or prevent disease transmission. A great variety of models have
been developed for this task, using different model structures,
covariates, and targets for prediction. Experience has shown that the
performance of these models varies; some tend to do better or worse in
different seasons or at different points within a season. Ensemble
methods combine multiple models to obtain a single prediction that
leverages the strengths of each model. We considered a range of ensemble
methods that each form a predictive density for a target of interest as
a weighted sum of the predictive densities from component models. In the
simplest case, equal weight is assigned to each component model; in the
most complex case, the weights vary with the region, prediction target,
week of the season when the predictions are made, a measure of component
model uncertainty, and recent observations of disease incidence. We
applied these methods to predict measures of influenza season timing and
severity in the United States, both at the national and regional levels,
using three component models. We trained the models on retrospective
predictions from 14 seasons (1997/1998 - 2010/2011) and evaluated each
model's prospective, out-of-sample performance in the five subsequent
influenza seasons. In this test phase, the ensemble methods showed
overall performance that was similar to the best of the component
models, but offered more consistent performance across seasons than the
component models. Ensemble methods offer the potential to deliver more
reliable predictions to public health decision makers.

\section{Introduction}\label{introduction}

The practice of combining predictions from different models has been
used for decades by climatologists and geophysical scientists. These
methods have subsequently been adapted and extended by statisticians and
computer scientists in diverse areas of scientific inquiry. In recent
years, these ``ensemble'' forecasting approaches frequently have been
among the top methods used in prediction challenges across a wide range
of applications.

Ensembles are a natural choice for noisy, complex, and interdependent
systems that evolve over time. In these settings, no one model is likely
to be able to capture and predict the full set of complex relationships
that drive future observations from a particular system of interest.
Instead ``specialist'' or ``component'' models can be relied on to
capture distinct features or signals from a system and, when combined,
represent a nearly complete range of possible outcomes. In this work, we
develop and compare a collection of ensemble methods for combining
predictive densities. This enables us to quantify the improvement in
predictions achieved by using ensemble methods with varying levels of
complexity.

To illustrate these ensemble methods, we present time-series forecasts
for infectious disease, specifically for influenza in the United States.
The international significance of emerging epidemic threats in recent
decades has highlighted the importance of understanding and being able
to predict infectious disease dynamics. With the revolution in science
driven by the promise of ``big'' and real-time data, there is an
increased focus on and hope for using statistics to inform public health
policy and decision-making in ways that could mitigate the impact of
future outbreaks. Some of the largest public health agencies in the
world, including the US Centers for Disease Control and Prevention (CDC)
have openly endorsed using models to inform decision making, saying
``with models, decision-makers can look to the future with confidence in
their ability to respond to outbreaks and public health emergencies''
\cite{cdc-decisions-2016}.

Development of the methods presented in this manuscript was motivated by
the observation that certain prediction models for infectious disease
consistently performed better than other models at certain times of
year. We observed in previous research that early in the U.S. influenza
season, simple models of historical incidence often outperformed more
standard time-series prediction models such as a seasonal
auto-regressive integrated moving average (SARIMA) model
\cite{ReichLabGitHubDiseasePredWithKCDEPackage}. However, in the middle
of the season, the time-series models showed improved accuracy. We set
out to determine whether ensemble methods could use this information
about past model performance to improve predictions.

A large number of ensemble methods have been developed for a diverse
array of tasks including regression, classification, and density
estimation. These methods are broadly similar in that they combine
results from multiple component models. However, details differ between
ensemble methods. We suggest Polikar \cite{polikar2006ensemble} for a
review of ensemble methods; many of these are also discussed in detail
in Hastie \etal \cite{Hastie2011}.

While there are many different methods for combining models, all
ensemble models discussed in this paper use an approach called stacking.
In this approach, each of the component models is trained separately in
a first stage, and cross-validated measures of performance of those
component models are obtained. Then, in a second stage, a stacking model
is trained using the cross-validated performance measures to learn how
to optimally combine predictive densities from the component models. The
specific implementations of stacking that we use obtain the final
predictive density as a weighted sum of the component predictive
densities, where the weights may depend on covariates. We refer to this
approach generally as a `'weighted density ensemble'' approach to
prediction. Several variations on this strategy have been explored in
the literature previously
\cite{smyth1999stackingDensityEstimators,  rigollet2007linearconvexaggregationdensity, ganti2011cake}.
However, other ensemble methods for density estimation have also been
developed. For example, Rosset and Segal \cite{rosset2002boosting}
develop a boosting method in which the component models are estimated
sequentially, with results from earlier models affecting estimation of
later models.

In structured prediction settings such as time series forecasting,
ensemble methods may benefit from taking advantage of the data
structure. For example, it may be the case that different models offer a
better representation of the data at different points in time. A common
idea in these settings is to use model weights that change over time.
For instance, model weights may vary as a function of how well each
model did in recent predictions \cite{herbster1998tracking} or by using
a more formal graphical structure such as a hidden Markov model to track
which component model is most likely to have generated new observations
as they arise over time
\cite{yamanishi2007dynamicmodelselection, cortes2014ensembleStructuredPrediction}.
It is also possible to combine the component models with weights that
depend on observed covariates or features \cite{Sill2009}. For example,
in an ensemble for a user recommendation system, Jahrer
\etal \cite{jahrer2010Netflix} allowed model weights to depend on a
variety of features including the time that a user submitted a rating.

Using component models that generate predictive densities for outcomes
of interest, we have implemented a series of ensembles using different
methods for choosing the weights for each model. Specifically, we
compare three different approaches. The first approach simply takes an
equally weighted average of all models. The second approach estimates
constant but not necessarily equal weights for each model. The third
approach is a novel method for determining model weights based on
features of the system at the time predictions are made. The overarching
goal of this study is to create a systematic comparison between ensemble
methods to study the benefits of increasing complexity in ensemble
weighting schemes.

We are aware of one previous article that has developed ensemble methods
for infectious disease prediction. Yamana \etal \cite{Yamana2016}
developed a model stacking framework that is similar to the second
approach outlined above using a constant weight for each component
model. The present article is differentiated from that work in that we
explore and compare a range of more flexible ensemble methods where the
weights depend on observed features.

This paper presents a novel ensemble method that determines optimal
model combinations based on (a) observed data at the time predictions
are made and (b) aspects of the predictive distributions obtained from
the component models. We refer to models built using this approach as
``feature-weighted'' ensembles. This approach fuses aspects of different
ensemble methods: it uses model stacking \cite{Wolpert1992} and
estimates model weights based on features of the system \cite{Sill2009}
using gradient tree boosting \cite{friedman2001greedy}.

Using seasonal influenza outbreaks in the US health regions as a
case-study, we developed and applied our ensemble models to predict
several attributes of the influenza season at each week during the
season. By illustrating the utility of these approaches to ensemble
forecasting in a setting with complex population dynamics, this work
highlights the importance of continued innovation in ensemble
methodology.

\section{Methods}\label{methods}

This paper presents a comparison of methods for determining weights for
weighted density ensembles, applied to forecasting specific features of
influenza seasons in the US. First, we present a description of the
influenza data we use in our application and the prediction targets.
Next, we discuss the three component models utilized by the ensemble
framework. We then turn to the ensemble framework itself, describing the
different ensemble model specifications used.

\subsection{Data and prediction
targets}\label{data-and-prediction-targets}

\begin{figure}[htbp]
\centering
\includegraphics{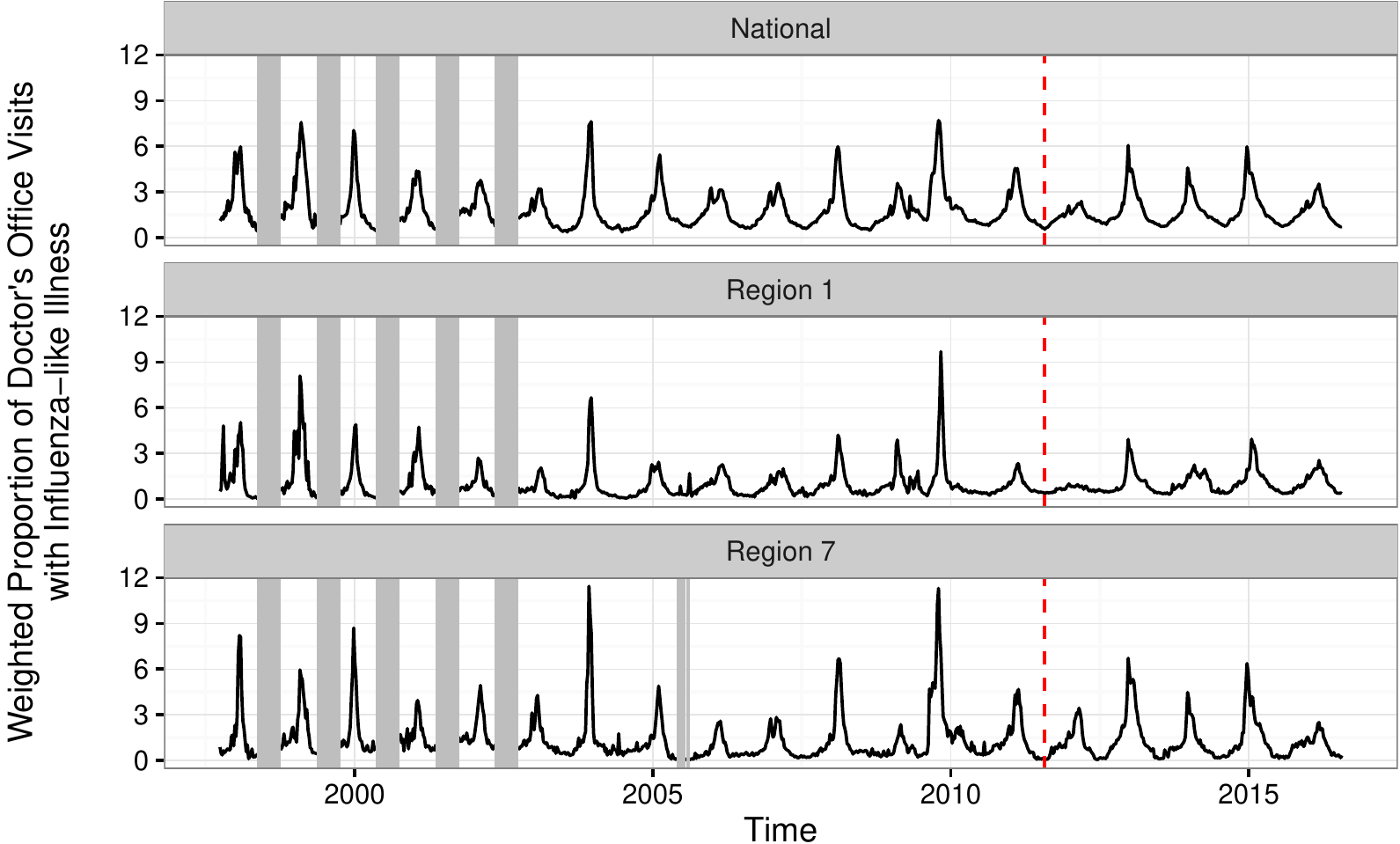}
\caption{\label{fig:raw-data}Plot of influenza data. The full data
include observations aggregated to the national level and for 10 smaller
regions. Here we plot only the data at the national level and in two of
the smaller regions; data for the other regions are qualitatively
similar. Missing data are indicated with vertical grey lines. The
vertical red dashed lines indicate the cutoff time between the training
and testing phases; 5 seasons of data were held out for testing.}
\end{figure}

We obtained publicly available data on seasonal influenza activity in
the United States between 1997 and 2016 from the U.S. Centers for
Disease Control and Prevention (CDC) (Fig \ref{fig:raw-data}). For each
of the 10 Health and Human Services regions in the country in addition
to the nation as a whole, the CDC calculates and publishes each week a
measure called the weighted influenza-like illness (wILI) index. The
wILI for a particular region is calculated as the average proportion of
doctor visits with influenza-like illness for each state in the region,
weighted by state population. During the CDC-defined influenza season
(between Morbidity and Mortality Weekly Report week 40 of one year and
20 of the next year), the CDC publishes updated influenza data on a
weekly basis. This includes ``current'' wILI data from two weeks prior
to the reporting date, as well as updates to previously reported numbers
as new data becomes available. For this analysis, we use only the final
reported wILI measures to train and predict from our models.

The CDC defines the influenza season onset as the first of three
successive weeks of the season for which wILI is greater than or equal
to a threshold that is specific to the region and season. This threshold
is the mean percent of patient visits where the patient had ILI during
low incidence weeks for that region in the past three seasons, plus two
standard deviations \cite{cdc2016}. The CDC provides historical
threshold values for each region going back to the 2007/2008 season
\cite{cdc2016-baselines}. Additionally, we define two other metrics
specific to a region-season. The peak incidence is the maximum observed
wILI measured in a season. The peak week is the week at which the
maximum wILI for the season is observed.

Each predictive distribution was represented by probabilities assigned
to bins associated with different possible outcomes. For onset week, the
bins are represented by integer values for each possible season week
plus a bin for ``no onset''. For peak week, the bins are represented by
integer values for each possible season week. For peak incidence, the
bins capture incidence rounded to a single decimal place, with a single
bin to capture all incidence over \(13.05\). Formally, the incidence
bins are as follows: {[}0, 0.05), {[}0.05, 0.15), \ldots{}, {[}12.95,
13.05), {[}13.05, \(\infty\)). These bins were used in the 2016-2017
influenza prediction contest run by the CDC
\cite{cdc2016-contest-guidelines}.

We measure the accuracy of predictive distributions using the log score.
The log score is a proper scoring rule \cite{Gneiting2007}, calculated
in our setting as the natural log of the probability assigned to the bin
containing the true observation. Proper scoring rules are preferred for
measuring the quality of predictive distributions because the expected
score is optimized by the true probabilty distribution. We note that for
peak week, in some region-seasons the same peak incidence was achieved
in multiple weeks (after rounding to one decimal place). In those cases,
we calculated the log score as the log of the sum of the probabilities
assigned to those weeks; this is consistent with scoring procedures used
in the 2016-2017 flu prediction contest run by the CDC
\cite{cdc2016-contest-guidelines}.

\subsection{Component models}\label{component-models}

We used three component models to generate probabilistic predictions of
the three prediction targets. The first model was a seasonal average
model that utilized kernel density estimation (KDE) to estimate a
predictive distribution for each target. The second model utilized
kernel conditional density estimation (KCDE) and copulas to create a
joint predictive distribution for incidence in all remaining weeks of
the season, conditional on recent observations of incidence
\cite{ReichLabGitHubDiseasePredWithKCDEPackage}. By calculating
appropriate integrals of this joint distribution, we constructed
predictive distributions for each of the seasonal targets. The third
model used a standard seasonal auto-regressive integrated moving average
(SARIMA) implementation. All models were fit independently on data
within each region.

\subsubsection{Kernel Density Estimation
(KDE)}\label{kernel-density-estimation-kde}

The simplest of the component models uses kernel density estimation
\cite{silverman1986density} to estimate a distribution for each target
based on observed values of that target in previous seasons within the
region of interest. We used Gaussian kernels and the default KDE
settings from the \texttt{density} function in the \texttt{stats}
package for R \cite{Rcore2015} to estimate the bandwidth parameter. For
the peak incidence target, we fit to log-transformed observations of
historical peak incidence. For the onset week prediction target, we
estimated the probability of no onset as the proportion of
region-seasons in all regions in the training phase where no week in the
season met the criteria for being a season onset.

To create an empirical predictive distribution of size \(N\) from a KDE
fit based on a data vector \(\by_{1:K}\) (for example, this might be the
vector of peak week values from the \(K\) training seasons), we first
drew \(N\) samples with replacement from \(\by_{1:K}\), yielding a new
vector \(\tilde \by_{1:N}\). We then drew a single psuedo-random deviate
from each of \(N\) truncated Gaussian distributions centered at
\(\tilde \by_{1:N}\) with the bandwidth estimated by the KDE algorithm.
The Gaussians we sampled from were truncated at the lower and upper
bounds of possible values for the given prediction target. Finally, we
discretized the sampled values to the target-specific bins. These
sampled points then make up the empirical predictive distribution from a
KDE model. We set the sample size to \(N = 10^5\). In theory, this model
assigns non-zero probability to every possible outcome; however, in a
few cases the empirical predictive distribution resulting from this
Monte Carlo sampling approach assigned probability zero to some of the
bins.

It is important to note that the predictions from this model do not
change as new data are observed over the course of the season.

\subsubsection{Kernel Conditional Density Estimation
(KCDE)}\label{kernel-conditional-density-estimation-kcde}

We used kernel conditional density estimation and copulas to estimate a
joint predictive distribution for flu incidence in each future week of
the season, and then calculated predictive distributions for each target
from that joint distribution
\cite{ReichLabGitHubDiseasePredWithKCDEPackage}. In our implementation,
we first used KCDE to obtain separate predictive densities for flu
incidence in each future week of the season. Each of these predictive
densities gives a conditional distribution for incidence at one future
time point given recent observations of incidence and the current week
of the season. KCDE can be viewed as a distribution-based analogue of
nearest-neighbors regression. We then used a copula to model dependence
among those individual predicitive densities, thereby obtaining a joint
predicitive density, or a distribution of incidence trajectories in all
future weeks.

To predict seasonal quantities (onset, peak timing, and peak incidence),
we simulate \(N = 10^5\) trajectories of disease incidence from this
joint predictive distribution. For each simulated incidence trajectory,
we compute the onset week, peak week, and peak incidence. We then
aggregate these values to create predictive distributions for each
target. This procedure for obtaining predictive distributions for the
targets of interest can be formally justified as an appropriate Monte
Carlo integral of the joint predictive distribution for disease
incidence in future weeks (see
\cite{ReichLabGitHubDiseasePredWithKCDEPackage} for details).

\subsubsection{Seasonal auto-regressive integrated moving average
(SARIMA)}\label{seasonal-auto-regressive-integrated-moving-average-sarima}

We fit seasonal ARIMA models \cite{Box2015} to wILI observations
transformed to be on the natural log scale. We manually performed
first-order seasonal differencing and used the stepwise procedure from
the \texttt{auto.arima} function in the \texttt{forecast} package
\cite{Hyndman2008} for R to select the specification of the
auto-regressive and moving average terms.

Similar to KCDE, forecasts were obtained by sampling \(N = 10^5\)
trajectories of wILI values over the rest of the season (using the
\texttt{simulate.Arima} function from the \texttt{forecast} package),
and predictive distributions of the targets were computed from these
sampled trajectories as described above.

\subsubsection{Component model training}\label{component-model-training}

We used data from 14 seasons (1997/1998 through 2010/2011) to train the
models. Data from five seasons (2011/2012 through 2015/2016) were held
out when fitting the models and used exclusively in the testing phase.
To avoid overfitting our models, we made predictions for the test phase
only once \cite{Hastie2011}.

Estimation of the ensemble models (discussed in the next subsection)
requires cross-validated measures of performance of each of the
component models in order to accurately gauge their relative
performance. For each region, we estimated the parameters of each
component model 15 times: 14 fits were obtained excluding one training
season at a time, and another fit used all of the training data. For
each fit obtained leaving one season out, we generated a set of three
predictive distributions (one for each of the prediction targets) at
each week in the held-out season. We were not able to generate
predictions from the SARIMA and KCDE models for some seasons in the
training phase because those models used lagged observations from
previous seasons that were missing in our data set. The component model
fits based on all of the training data were used to generate predictions
for the test phase.

\subsection{Ensemble models}\label{ensemble-models}

All of the ensemble models we consider in this article work by averaging
predictions from the component models to obtain the ensemble prediction.
Additionally, these methods are stacked model ensembles because they use
leave-one-season-out predictions from the independently estimated
component models as inputs to estimate the model weights
\cite{Wolpert1992}. We begin our discussion of ensemble methods with a
general overview, introducing a common set of notation and giving a
broad outline of the ensemble models we will use in this article. We
then describe our proposed weighted density ensemble model
specifications in more detail.

\subsubsection{Overview of ensemble
models}\label{overview-of-ensemble-models}

A single set of notation can be used to describe all of the ensemble
frameworks implemented here. Let \(f_m(y_t|\bx_t^{(m)})\) denote the
predictive density from component model \(m\) for the value of the
scalar random variable \(Y_t\) conditional on observed variables
\(\bx_t^{(m)}\). Observations of disease incidence are reported weekly
in our data set, so \(t\) indexes the week of the season. The variable
\(Y_t\) could for example represent the peak incidence for a given
season and region; in our application to predicting seasonal quantities,
the same outcome \(y_t\) will be realized for all weeks within a given
season. In the context of time series predictions, the covariate vector
\(\bx_t^{(m)}\) may include time-varying covariates such as the week at
which the prediction is made or lagged incidence. The superscript
\(^{(m)}\) reflects the fact that each component model may use a
different set of covariates.

The combined predictive density \(f(y_t|\bx_t)\) for a particular target
can be written as

\begin{equation}
f(y_t|\bx_t) = \sum_{m = 1}^M \pi_m(\bx_t) f_m(y_t|\bx_t^{(m)}). \label{eqn:EnsembleModel} 
\end{equation}

\begin{figure}[htbp]
\centering
\includegraphics{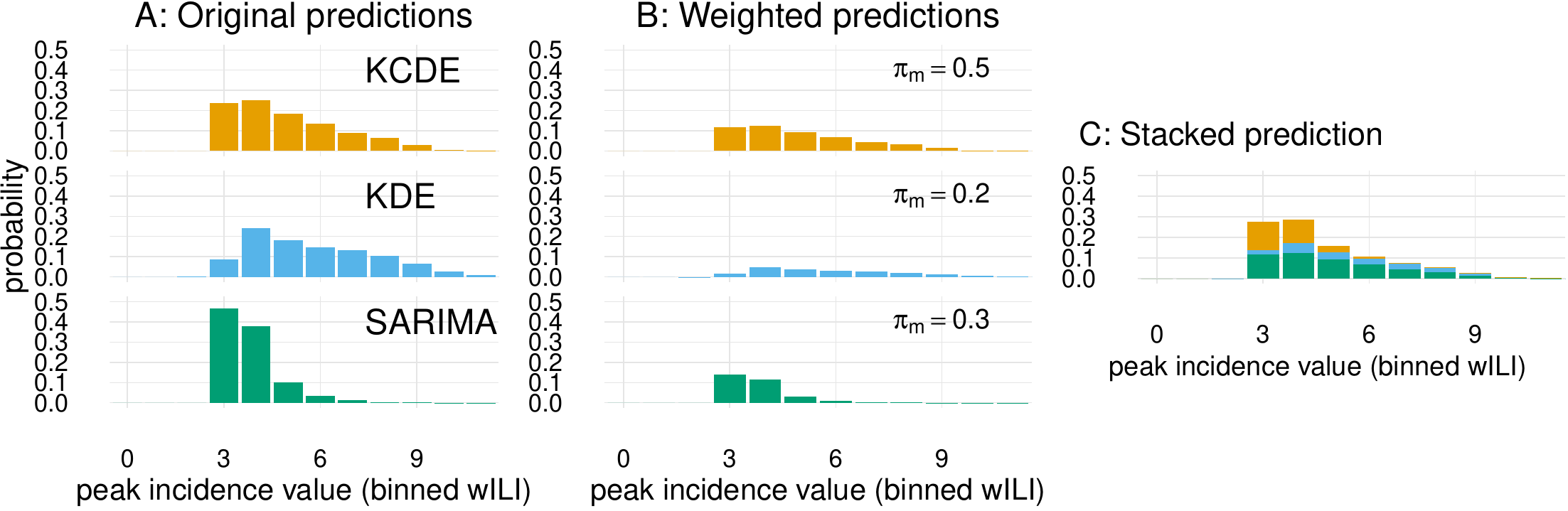}
\caption{\label{fig:stacking-concept}Conceptual diagram of how the
stacking models operate on probabilistic predictive distributions. The
distributions illustrated here have density bins of 1 wILI unit, which
differs from those used in the manuscript for illustrative purposes
only. Panel A shows the predictive distributions from three component
models. Panel B shows scaled versions of the distributions from A, after
being multiplied by model weights. In Panel C, the scaled distributions
are literally stacked to create the final ensemble predictive
distribution.}
\end{figure}

In Equation \eqref{eqn:EnsembleModel} the \(\pi_m\) are the model
weights, which are allowed to vary as a function of observed features in
\(\bx_t\). We define \(\bx_t\) to be a vector of all observed quantities
that are used by any of the component models or in calculating the model
weights. In order to guarantee that \(f(y_t|\bx_t)\) is a probability
distribution we require that \(\sum_{m = 1}^M \pi_m(\bx_t) = 1\) for all
\(\bx_t\). Fig \ref{fig:stacking-concept} illustrates the concept of
stacking the predictive densities for each component model.

In the following subsection, we propose a framework for estimating
\emph{feature-dependent weights} for a stacked ensemble model. By
\emph{feature-dependent} we mean that the weights associated with
different component models are driven by observed features or
covariates. Although we illustrate the method in the context of
time-series predictions, the method could be used in any setting where
we wish to combine distribution estimates from multiple models. Features
could include observed data from the system being predicted (such as
recent wILI measurements or the time of year at which predictions are
being made), observed data from outside the system (for example, recent
weather observations), or features of the predictions themselves
(e.g.~summaries of the predictive distributions from the component
models, such as a measure of spread in the distribution, or the time
until a predicted peak). Based on exploration of training phase data and
\emph{a priori} knowledge of the disease system, we chose three features
of the system to illustrate the proposed ``feature-weighting''
methodology: week of season, component model uncertainty (defined as the
minimum number of predictive distribution bins required to cover 90\%
probability), and wILI measurement at the time of prediction. These
features were chosen prior to and not changed after implementing
test-phase predictions.

We used four distinct methodologies to define weights to use for the
stacking models:

\begin{enumerate}
\def\labelenumi{\arabic{enumi}.}
\item
  Equal Weights (\textbf{EW}): \(\pi_m(\bx_t) = 1/M\). In this scenario,
  each model contributes the same weight for each target and for all
  values of \(\bx_t\).
\item
  Constant model weights via degenerate EM (\textbf{dEM}):
  \(\pi_m(\bx_t) = c_m\), a constant where \(\sum_{m=1}^M c_m = 1\) but
  the constants are not necessarily the same for each model. These
  weights are estimated using the degenerate estimation-maximization
  (dEM) algorithm \cite{Lin2004}. A separate set of weights is estimated
  for each region and prediction target.
\item
  Feature-weighted (\textbf{FW}): \(\pi_m(\bx_t)\) depends on features
  including week of the season and model uncertainty for the KCDE and
  SARIMA models. A separate set of weighting functions is estimated for
  each region and prediction target.
\item
  Feature-weighted with regularization: \(\pi_m(\bx_t)\) depends on
  features, but with regularization discouraging the weights from taking
  extreme values or from varying too quickly as a function of \(\bx_t\).
  A separate set of weighting functions is estimated for each region and
  prediction target. We fit three variations on this ensemble model,
  using different sets of features:

  \begin{enumerate}
  \def\labelenumii{\alph{enumii}.}
  \tightlist
  \item
    (\textbf{FW-reg-w}) week of the season;
  \item
    (\textbf{FW-reg-wu}) week of the season and model uncertainty for
    the KCDE and SARIMA models;
  \item
    (\textbf{FW-reg-wui}) week of the season, model uncertainty for the
    KCDE and SARIMA models, and incidence (wILI) in the most recent
    week.
  \end{enumerate}
\end{enumerate}

All in all, this leads to 6 ensemble models, summarized in Table
\ref{tbl:EnsembleModelSummaryTable}. The first three of these models
(\textbf{EW}, \textbf{dEM}, and \textbf{FW}) can be viewed as variations
on \textbf{FW-reg-wu} if we vary the amount and type of regularization
imposed on the \textbf{FW-reg-wu} model. Thus, comparisons among these
four models will enable us to explore the benefits of allowing the model
weights to depend on covariates while imposing an appropriate amount of
rigidity on the model weight functions \(\pi_m(\bx_t)\). We will discuss
the regularization strategies used in \textbf{FW-reg-wu} further in the
next subsection. Meanwhile, comparisons among the \textbf{FW-reg-w},
\textbf{FW-reg-wu}, and \textbf{FW-reg-wui} models will allow us to
explore the relative contributions to predictive performance that can be
achieved by allowing the model weights to depend on different features.

\begin{table}[!ht]
\centering
\caption{\label{tbl:EnsembleModelSummaryTable}Summary of ensemble methods and what the model weights depend on.}
\begin{tabular}{rcccccc}
\toprule
         & \multicolumn{6}{c}{Component Model Weights Vary with...} \\
\cline{2-7}
  &   & Prediction & Week of & SARIMA & KCDE & Current \\ 
Model & Region & Target & Season & Uncertainty & Uncertainty & wILI \\ 
  \hline
EW         &   &   &   &   &   &   \\
CW        & X & X &   &   &   &   \\
FW         & X & X & X & X & X &   \\
FW-reg-w   & X & X & X &   &   &   \\
FW-reg-wu  & X & X & X & X & X &   \\
FW-reg-wui & X & X & X & X & X & X \\
\bottomrule
\end{tabular}
\end{table}

Each of the six ensemble models, along with the three component models,
are used to generate predictions in every season-week of each of the
five testing seasons, assuming perfect reporting. These predictions are
then used to evaluate the prospective predictive performance of each of
the ensemble methods. In total, we evaluate 9 models in 11 regions over
5 years and 3 targets of interest.

\subsection{Feature-weighted stacking
framework}\label{feature-weighted-stacking-framework}

In this section we introduce the particular specification of the
parameter weight functions \(\pi_m(\bx_t)\) that we use for the
\textbf{FW}, \textbf{FW-reg-w}, \textbf{FW-reg-wu}, and
\textbf{FW-reg-wui} models and discuss estimation.

In order to ensure that the the \(\pi_m\) are non-negative and sum to 1
for all values of \(\bx_t\), we parameterize them in terms of the
softmax transformation of real-valued latent functions \(\rho_m\)

\begin{equation}
\pi_{m}(\bx_t) = \frac{\exp\{\rho_m(\bx_t)\}}{\sum_{m' = 1}^M \exp\{\rho_{m'}(\bx_t)\}}.  \label{eqn:PiSoftmaxRho} 
\end{equation}

For a pair of models \(l, m \in \{1, ..., M\}\),
\(\rho_l(\bx_t) > \rho_m(\bx_t)\) indicates that model \(l\) has more
weight than model \(m\) for predictions at the given value of \(\bx_t\).
The functions \(\rho_m(\bx_t)\) could be parameterized and estimated
using many different techniques, such as a linear specification in the
features, splines, or so on. We chose to estimate the functions
\(\rho_m(\bx)\) using gradient tree boosting.

Gradient tree boosting uses a forward stagewise additive modeling
algorithm to iteratively and incrementally construct a series of
regression trees that, when added together, create a function designed
to minimize a given loss function. In our application, the algorithm
builds up the \(\rho_m(\bx_t)\) that minimize the negative log-score of
the stacked predictions \(f(y_t|\bx_t)\) across all times \(t\):

\begin{align}
L\{\brho(\bx_t)\} &= - \sum_{t} \log\{f(y_t|\bx_t)\} \nonumber \\
&= - \sum_{t} \log\left[\sum_{m = 1}^M \frac{\exp\{\rho_m(\bx_t)\}}{\sum_{m' = 1}^M \exp\{\rho_{m'}(\bx_t)\}}f_m(y_t|\bx_t^{(m)})\right], \label{eqn:logloss}
\end{align}

where \(f_m(y_t|\bx_t^{(m)})\) is the cross-validated predictive density
from the \(m\)th model evaluated at the realized outcome \(y_t\).

Specifically, we define a single tree as

\begin{equation}
T(\bx_t; \btheta) = \sum_{j=1}^J \gamma_j I_{R_j}(\bx_t),
\end{equation}

where the \(R_j\) are a set of disjoint regions that comprise a
partition of the space \(\mathcal{X}\) of feature values \(\bx_t\), and
\(I\) is the indicator function taking the value \(1\) if
\(\bx_t \in R_j\) and \(0\) otherwise. The parameters
\(\btheta = (\bpsi, \bgamma)\) for the tree are the split points
\(\bpsi\) partitioning \(\mathcal{X}\) into the regions \(R_j\) and the
regression constants \(\bgamma\) associated with each region. The
function \(\rho_m(\bx_t)\) is obtained as the sum of \(B\) trees:

\begin{equation}
\rho_m(\bx_t; \Theta_m) = \sum_{b=1}^B T(\bx_t; \btheta_{m, b}).
\end{equation}

In each iteration \(b\) of the boosting process, we estimate \(M\) new
regression trees, one for each component model. These trees are
estimated so as to minimize a local approximation to the loss function
around the weight functions that were obtained after the previous
boosting iteration. Our approach builds on the \texttt{xgb.train}
function in the \texttt{xgboost} package for \texttt{R} to perform this
estimation \cite{xgboost}. The functionality in that package assumes
that the loss function is convex, and optimizes a quadratic
approximation to the loss in each boosting iteration. The loss function
in Equation \eqref{eqn:logloss} is not guaranteed to be convex, so a
direct application of this optimization method fails in our setting. We
have modified the implementation in the \texttt{xgboost} package to use
a gradient descent step in cases where the loss is locally concave.

Gradient tree boosting is appealing as a method for estimating the
functions \(\rho_m\) because it offers a great deal of flexibility in
how the weights can vary as a function of the features \(\bx_t\). On the
other hand, this flexibility can lead to overfitting the training data.
In order to limit the chances of overfitting, we have explored the use
of three regularization parameters:

\begin{enumerate}
\def\labelenumi{\arabic{enumi}.}
\item
  The number of boosting iterations \(B\). As \(B\) increases, more
  extreme weights (close to 0 or 1) and more rapid changes in the
  weights as \(\bx\) varies are possible.
\item
  An \(L_1\) penalty on the number of tree leaves, \(J\). A large
  penalty encourages the regression trees to have fewer leaves, so that
  there is less flexibility for the model weights to vary as a function
  of \(\bx_t\).
\item
  An \(L_1\) penalty on the regression constants \(\gamma_j\). A large
  penalty encourages these constants to be small, so that the overall
  model weights change less in each boosting iteration.
\end{enumerate}

We selected values for these regularization parameters using a grid
search optimizing leave-one-season-out cross-validated model
performance.

\subsection{Software and code}\label{software-and-code}

We used R version 3.2.2 (2015-08-14) for all analyses \cite{Rcore2015}.
All data and code used for this analysis is freely available in an R
package online at
\url{https://github.com/reichlab/adaptively-weighted-ensemble} and may
be installed in R directly. Predictions generated in real-time with
early development versions of this model during the 2016/2017 influenza
season may be viewed at \url{https://reichlab.io/flusight/}. To maximize
reproducibility of our work, we have set seeds prior to running code
that relies on stochastic simulations using the \texttt{rstream} package
\cite{Leydold2015}. Additionally, the manuscript itself was dynamically
generated using RMarkdown.

\section{Results}\label{results}

To evaluate overall model performance, we computed log scores for all
predictions made by each model across all regions and test phase
seasons. We also examined results for predictions made before the peak
week (for predictions of peak timing or peak incidence) or the season
onset (for predicitons of onset timing) within each of the test phase
seasons.

\subsubsection{Feature-weighted ensemble model weights reflect trends in
component model log
scores}\label{feature-weighted-ensemble-model-weights-reflect-trends-in-component-model-log-scores}

\begin{figure}[htbp]
\centering
\includegraphics{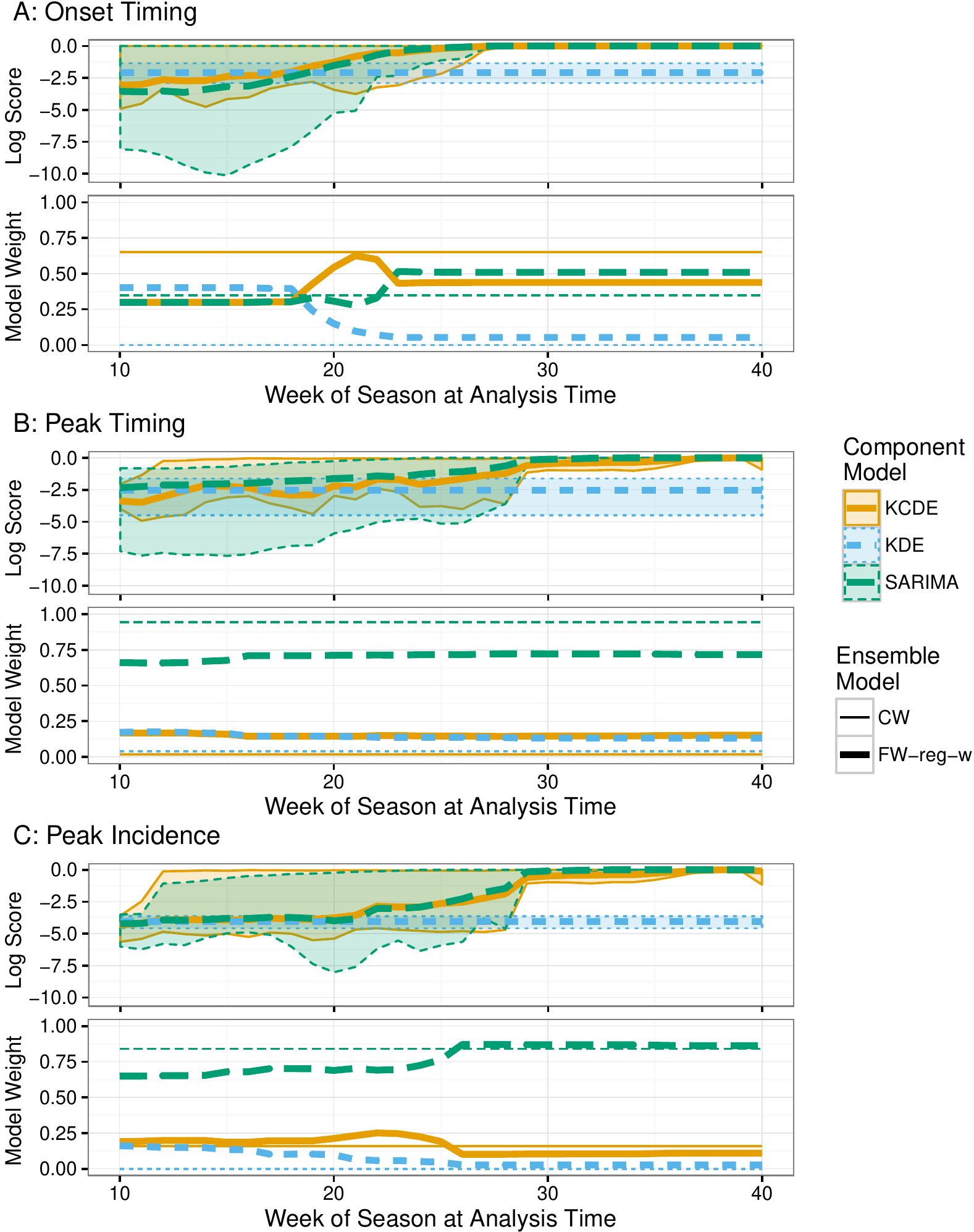}
\caption{\label{fig:example-weights}Example of component model weights
from the \textbf{CW} and \textbf{FW-reg-w} models for National
predictions. The upper plot within each panel shows mean, minimum, and
maximum log scores achieved by each component model for predictions of
the given prediction target at the national level in each week of the
season, summarizing across all seasons in the training phase when all
three component models produced predictions. The lower plot within each
panel shows model weights from the \textbf{CW} and \textbf{FW-reg-w}
ensemble methods at each week in the season.}
\end{figure}

Fig \ref{fig:example-weights} displays variation in leave-one-season-out
log scores from the three component models over the course of the
training phase seasons, along with the corresponding model weight
estimates from the \textbf{CW} and \textbf{FW-reg-w} models. Performance
of the \textbf{SARIMA} and \textbf{KCDE} models is similar, with mean
log scores from those models starting out near or slightly below the
mean performance of \textbf{KDE}, but with performance improving as more
data become available. Near the beginning of some seasons, predictions
from the \textbf{SARIMA} model are quite a bit worse than predictions
from the other two component models. Supplemental Fig 1 illustrates that
these patterns are consistent across the other regions. Supplemental Fig
2 shows that performance of the component models also varies with the
model's uncertainty as measured by the number of bins required to cover
90\% in the predictive distribution, and Supplemental Fig 3 shows that
performance varies with the observed wILI in the week when predictions
are made.

\begin{figure}[htbp]
\centering
\includegraphics{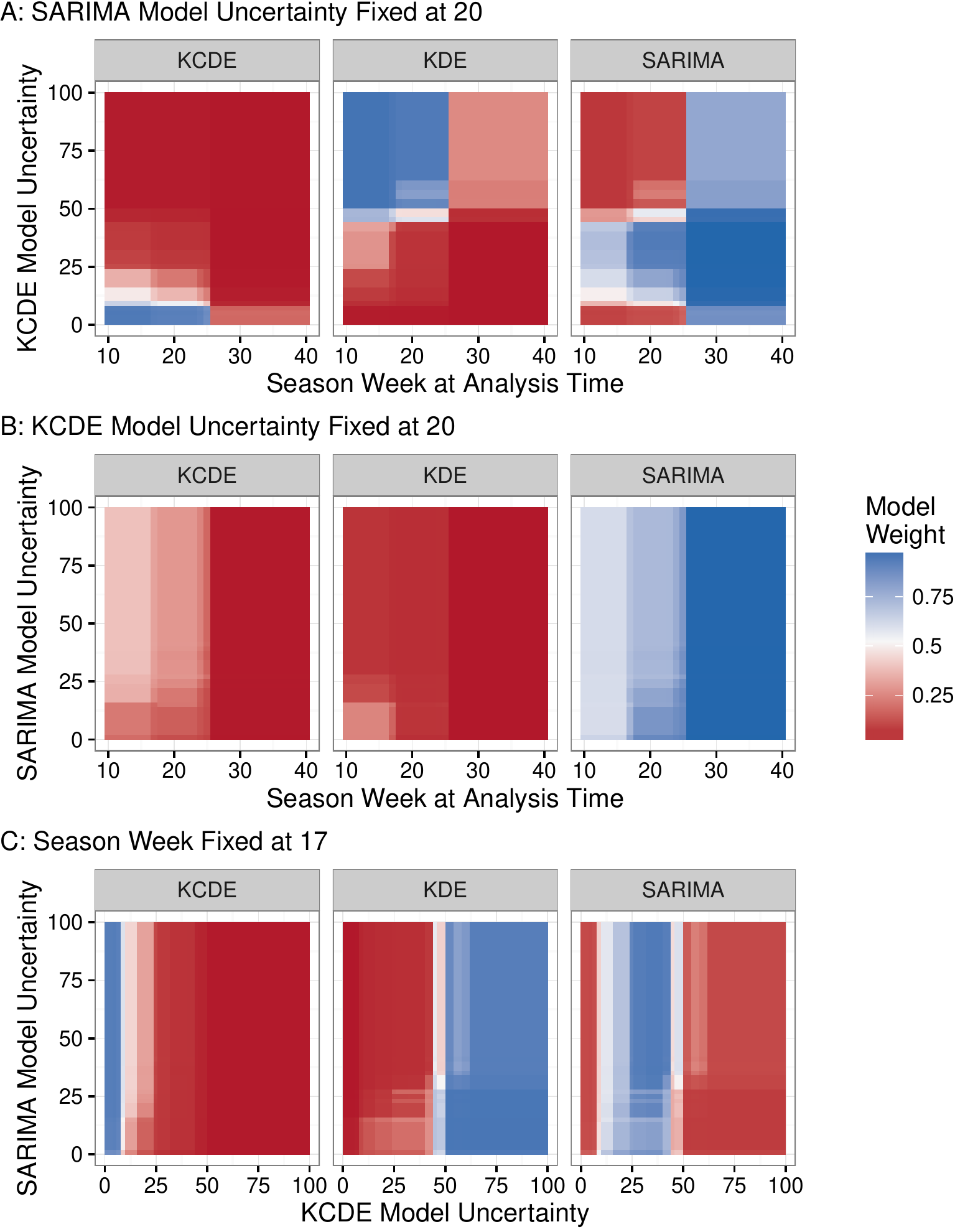}
\caption{\label{fig:FWregwuModelWeights}Weights assigned to each
component model by the FW-reg-wu model for the prediction of season peak
incidence at the national level. There are three weighting functions
(one for each component model) represented in each row of the figure.
The value of the weight is depicted by the color. Each function depends
on three features: the week of the season at the time when the
predictions are made, KCDE model uncertainty, and SARIMA model
uncertainty. Model uncertainty represents the minimum number of
predictive distribution bins required to cover 90\% probability of the
predictive distribution, so the higher this number is the more uncertain
the model is.}
\end{figure}

The model weights assigned by the feature weighted ensemble models
generally track these trends in relative model performance (Figs
\ref{fig:example-weights}, \ref{fig:FWregwuModelWeights}). For all three
targets, at the national level the weight assigned to the
\textbf{SARIMA} model increases and the weight assigned to \textbf{KDE}
decreases as the season progresses. However, the magnitude of shifts in
model weights as the weighting features vary is different for the three
prediction targets.

\subsubsection{Best models have similar aggregate
performance}\label{best-models-have-similar-aggregate-performance}

\begin{figure}[htbp]
\centering
\includegraphics{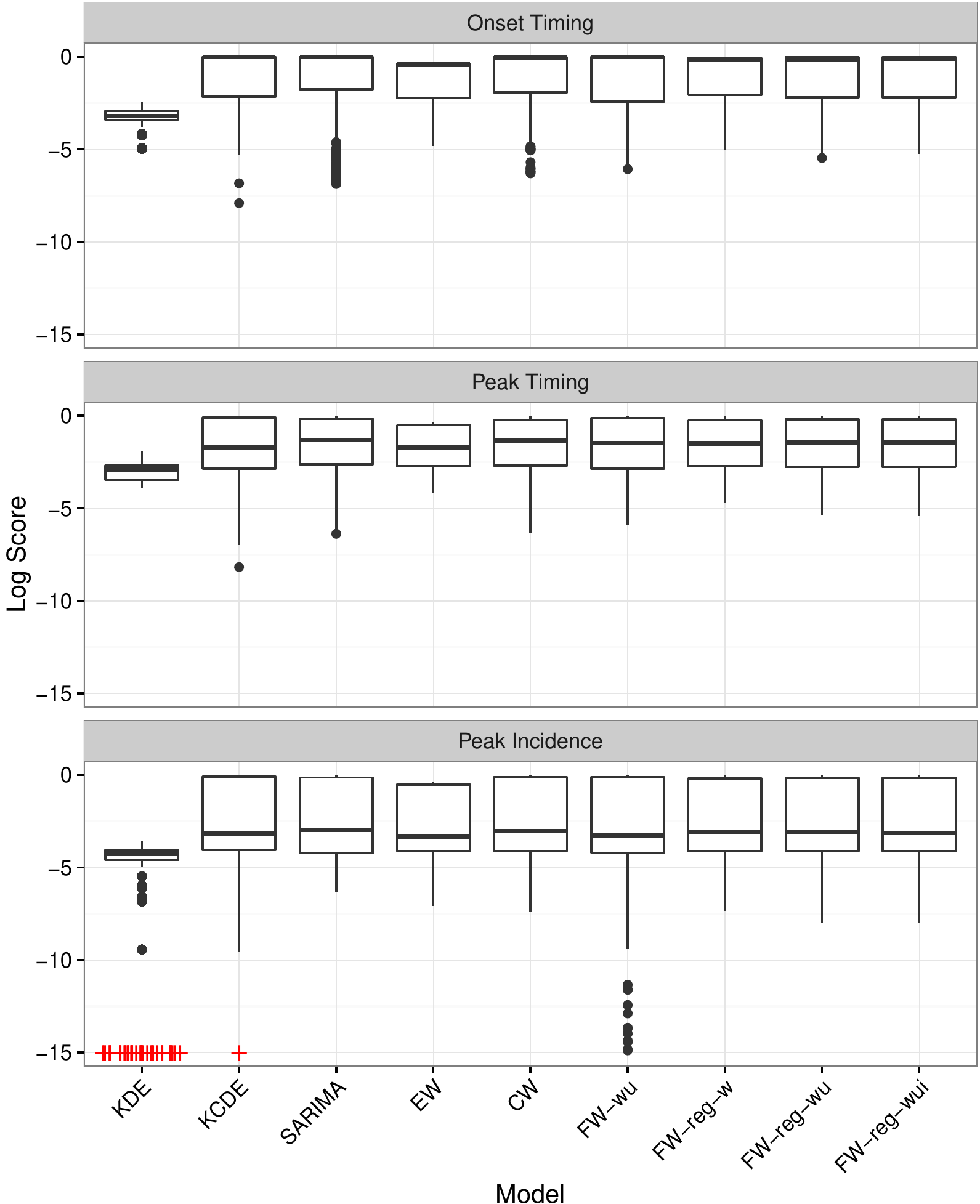}
\caption{\label{fig:test_phase_log_score_boxplot}Log scores across all
regions, seasons, and season weeks represented in box plots. Log scores
of negative infinity are represented with a cross at -15.}
\end{figure}

Aggregating across all combinations of region, season, and week of the
season in the test phase, most of the models had similar performance
(Fig \ref{fig:test_phase_log_score_boxplot}). The most important
exception to this is the \textbf{KDE} model, which achieved consistently
lower log scores than the other methods, including several cases where
the Monte Carlo sampling procedure we used to approximate the predictive
distribution assigned probability 0 to the true peak incidence bin
(there was one such case for the \textbf{KCDE} model). The low
performance of the \textbf{KDE} model pulled the log scores for the
\textbf{EW} method slightly below the other methods, and resulted in
some outlying cases where the unregularized \textbf{FW-wu} method
performed poorly for predicting peak incidence. However, aggregated
performance of the \textbf{KCDE}, \textbf{SARIMA}, \textbf{CW}, and
variations on the \textbf{FW} models was quite similar in most cases. A
linear mixed effects model was not able to distinguish any statistically
significant differences in mean performance of these methods
(Supplemental Figs 4 and 5).

\subsubsection{Ensembles show stable performance across seasons for
early-season
predictions}\label{ensembles-show-stable-performance-across-seasons-for-early-season-predictions}

\begin{figure}[htbp]
\centering
\includegraphics{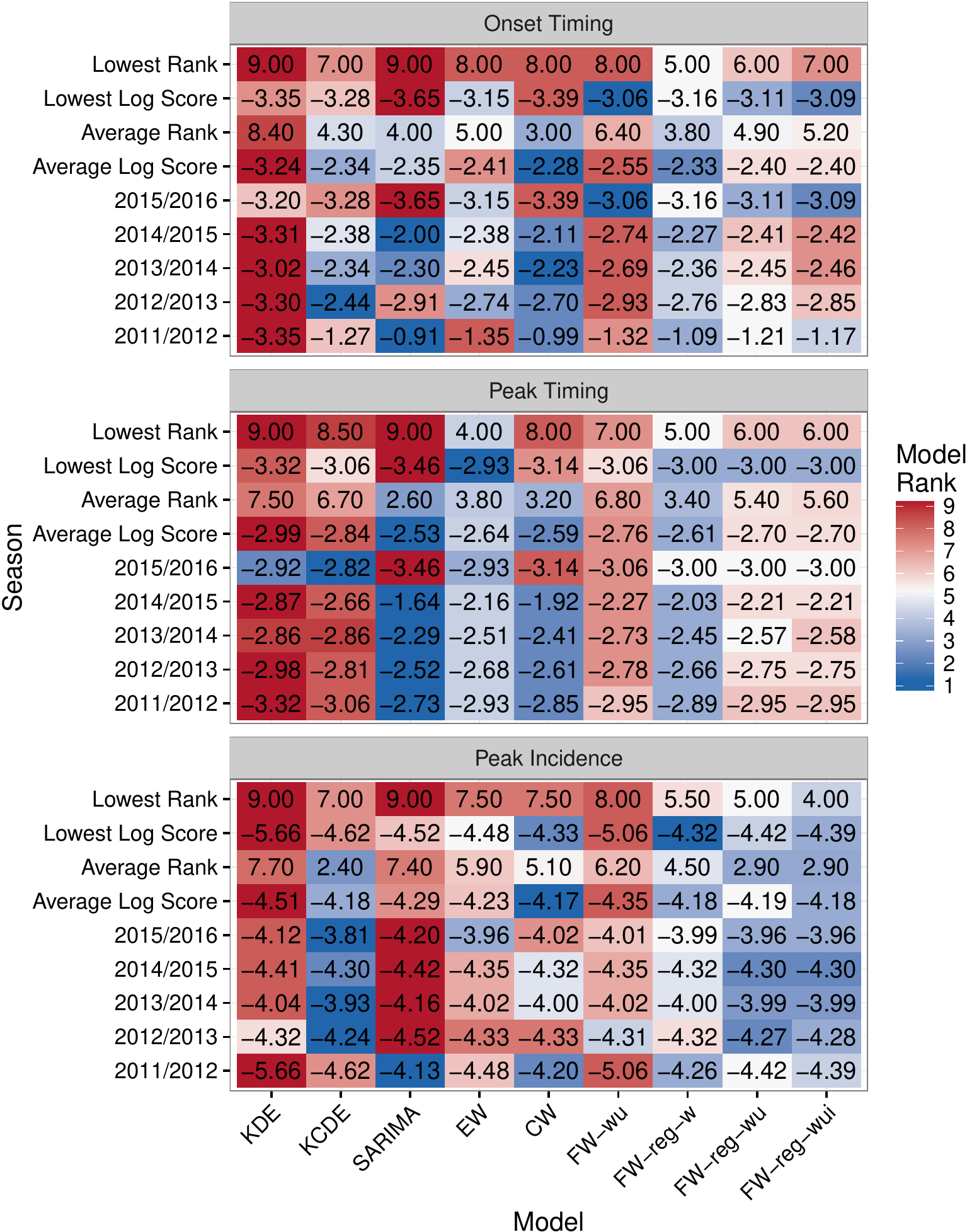}
\caption{\label{fig:test_phase_log_scores_heatmap}Model performance
ranked by mean log score within each of the five test seasons for
predictions made before the target (season onset or peak) occurred.
Averages are taken across all regions.}
\end{figure}

Although the aggregate performance of these models is quite similar,
some differences between the methods begin to emerge when we examine
performance in more detail. Predictions made before the season peak (for
predictions of peak incidence or peak timing) or before the season onset
(for predictions of season onset timing) are the most relevant to
decision makers using the predictions as inputs to set public policy.
Additionally, it is important that these predictions be of consistent
quality in all seasons, whether the seasonal dynamics follow historic
seasonal trends or diverge from those common patterns. We evaluated the
relative performance of predictions from each model that were generated
before the onset or peak week in each test season (Fig
\ref{fig:test_phase_log_scores_heatmap}).

In all test phase seasons and for all prediction targets, the worst
performing model is always one of the three component models
(\textbf{KDE}, \textbf{KCDE}, or \textbf{SARIMA}). Furthermore, each of
the component models outperforms the other two for at least one
combination of season and prediction target. Within each season, the
component models are often among either the best-performing or the
worst-performing models. For example, the \textbf{SARIMA} model ranks
first for predictions of peak timing in four out of five seasons and
last in the fifth season, but ranks last for predictions of peak
incidence in four out of five seasons and first in the remaining season.
Similar trends, though slightly less extreme, also hold for \textbf{KDE}
and \textbf{KCDE}.

The ensemble methods have more stable performance across the test phase
seasons. Only \textbf{FW-reg-w} was among the top half of models in all
seasons for predictions of onset timing; only \textbf{EW} and
\textbf{FW-reg-w} were among the top half of models in all seasons for
predictions of peak timing; and only \textbf{FW-reg-wu} and
\textbf{FW-reg-wui} were among the top half of models in all seasons for
predictions of peak incidence (Fig 5)

Model consistency can also be measured with the minimum of the log
scores achieved across the test phase seasons. Here again the ensemble
methods outperform the component models. In particular, worst-case
performance of the \textbf{EW} and regularized \textbf{FW} methods is in
the top half of all methods for all three prediction targets (Fig 5).

For example, consider predictions of peak timing, which could be used to
plan the deployment of medical resources \cite{cdc-decisions-2016}.
Among the component models we considered, \textbf{SARIMA} had the
highest mean log score across the test phase, and was the best model in
four out of five test-phase seasons. However, it had the worst
performance of all models we considered in the 2015/2016 season. The
\textbf{EW} and \textbf{FW-reg-w} ensembles achieved only slightly lower
average log scores than \textbf{SARIMA} overall, were among the
best-performing methods in all five test seasons, and assigned an
average of about 60 to 70 percent more probability to the eventually
realized peak week than \textbf{SARIMA} in predictions made during the
season that \textbf{SARIMA} struggled.

\subsubsection{Regularization improves feature-weighted ensemble
models}\label{regularization-improves-feature-weighted-ensemble-models}

The regularization of feature-weighted ensembles improved early-season
prediction accuracy for all three metrics. Specifically, a comparison of
average log-scores for the \textbf{FW-wu} and \textbf{FW-reg-wu} models
show consistent improvement in the model that used regularization to
create smoother functions of model weights as a function of season week
and model uncertainty. Only once out of 15 combinations of target and
test-season (onset timing in the 2015/2016 season) did \textbf{FW-wu}
outperform its regularized counterpart (Fig 5).

\section{Discussion}\label{discussion}

In this work we have examined the potential for ensemble methods to
improve infectious disease predictions. We explored a nested series of
ensemble methods, focusing on methods that computed weighted averages of
predictive distributions for seasonal targets of public health interest,
such as the peak intensity of the outbreak and the timing of both season
onset and peak. The methods we examined ranged from using equal model
weights to more complex schemes with weights that varied as functions of
multiple covariates. These ensemble methods achieved overall performance
that was about as good as the individual component models, with
increased stability in model performance across different seasons.

Increased stability in predictive accuracy can provide decision makers
with more confidence when using predictions as inputs to set policy. For
example, if a single model does well in most seasons but occasionally
fails badly, planning decisions may be negatively impacted in those
failing years. This may be particularly important in a public health
setting where the events that are most important to get right are those
relatively rare cases when incidence is much larger than usual or the
season timing is earlier or later than usual. This reduction in
variability of model performance achieved by ensemble methods is
therefore important for ensuring that our predictions are reliable under
a variety of conditions.

The different ensemble specifications we considered had similar average
performance over the test phase, when all models were making
prospective, out-of-sample predictions. However, there were differences
among the ensemble models in terms of their consistency across different
seasons. For all three prediction targets, worst-case log scores and the
worst-case model ranking across the five test seasons were slightly
lower for the constant-weight (\textbf{CW}) ensemble than they were for
the ensemble with smoothed model weights varying by week
(\textbf{FW-reg-w}). There were not appreciable or consistent
differences among the feature-weighted models using different feature
sets, indicating that including model uncertainty and recent
observations of disease incidence did not add much more information
about relative model performance than was available from the week of the
season in which predictions were generated. The equally weighted
ensemble model had lower mean log scores than the \textbf{CW} and
\textbf{FW-reg-w} ensembles for all three prediction targets, but had
better worst-case performance than the other ensembles for predictions
of peak timing. Synthesizing these observations, no ensemble was
uniformly better than the others, but the \textbf{FW-reg-w} method had
good average and worst-case performance across all test phase seasons
and prediction targets.

The feature-weighted ensemble models presented in this article use a
novel scheme to estimate feature-dependent model weights that sum to 1
and are therefore suitable for use in combining predictive
distributions. This general method could be applied to combine
distribution estimates in any context, and is not limited to time-series
or infectious disease applications. Furthermore, comparing an
implementation of the feature-weighting that smoothed the model weights
to one that did not showed consistent improvements in model performance.
This result suggests that future work on feature-weighted ensemble
implementations should consider regularized estimation.

Infectious disease predictions are only useful to public health
officials if they are communicated effectively in real time. Predictions
from an early version of the \textbf{FW-reg-w} model were updated weekly
during the 2016/2017 influenza season and disseminated through an
interactive website at \url{https://reichlab.io/flusight/}.

A central challenge of working with infectious disease data sets is the
limited number of years of data available for model estimation and
evaluation. We have used approximately one fourth of our data set for
model evaluation, which left us with only 14 seasons of training data
and 5 seasons of testing data. Additionally, we had fewer than 14
seasons of leave-one-season-out predictions to use in estimating the
model weighting functions for the \textbf{FW} ensemble methods because
the \textbf{SARIMA} model required unobserved seasonally lagged
incidence to make predictions for the first few seasons in the training
phase. This small sample size may have negatively impacted our ability
to estimate the weighting functions. We also have a small effective
sample size for detecting differences in average model performance in
the test phase because of the high degree of correlation in model log
scores for the same prediction target in different weeks and regions
within the same season.

Another limitation of this work is the small selection of component
models used. Theoretical results and applications have demonstrated that
ensemble methods are most effective when using a diverse set of
component models \cite{polikar2006ensemble}. In our study, the
\textbf{KCDE} and \textbf{SARIMA} component models are similar in that
they both use seasonal terms and observations of recent incidence to
(though we note that these two models tended to perform well in
different seasons, as illustrated in Fig
\ref{fig:test_phase_log_scores_heatmap}). Increased component model
diversity could yield improved ensemble performance; this could be
achieved either through inclusion of different model structures (such as
agent-based or mechanistic models) or different covariates (such as
spatial effects, weather, or circulating strains of a disease).

Our exploration of feature-weighted ensembles is also limited by the
relatively restricted feature sets we used for the weighting functions.
We selected a few features based on exploratory analysis of the training
phase results, and set all ensemble model formulations before obtaining
any predictions for the test phase. It is possible that other weighting
features not considered in this work may be more informative than those
we have used. Some ideas for weighting covariates to use in future work
include the largest incidence so far this season; the onset threshold;
alternative summaries of the predictive distributions from the component
models such as the probability at the mode or the modal value; the
predominant flu strain; or the distribution of incidence in age groups.

This work provides a rigorous and comprehensive evaluation of ensemble
methods for averaging probabilistic predictions for features of
infectious disease outbreaks. A range of models, both single component
models and ensemble models that combined component model predictions,
demonstrated the ability to make more accurate predictions than a
seasonal average baseline model. Additionally, systematic comparisons of
simple and complex prediction models highlight a crucial added value of
ensemble modeling, namely increased stability and consistency of model
performance across seasons. Continued investigation, application, and
innovation is necessary to strengthen our understanding of how to best
leverage combinations of models to assist decision makers in fields,
such as public health and infectious disease surveillance, that require
data-driven rapid response.

\section{Acknowledgments}\label{acknowledgments}

This work was supported by Award Number R35GM119582 from the National
Institute Of General Medical Sciences and Defense Advanced Projects
Research Agency Young Faculty Award Number Dl6AP00144. The content is
solely the responsibility of the authors and does not necessarily
represent the official views of the National Institute Of General
Medical Sciences, the National Institutes of Health, or the Defense
Advanced Projects Research Agency. The funders had no role in study
design, data collection and analysis, decision to publish, or
preparation of the manuscript.

\bibliographystyle{plos2015}\bibliography{feature-weighted-ensembles}

\hypertarget{refs}{}

\end{document}